\documentclass[letterpaper]{article}
\usepackage{aaai}
\usepackage{times}
\usepackage{helvet}
\usepackage{courier}
\usepackage{natbib}
\usepackage{mathtools}
\usepackage{amsmath}    
\usepackage{multirow}
\usepackage{booktabs}
\usepackage{listings}
\usepackage{color}

\definecolor{dkgreen}{rgb}{0,0.6,0}
\definecolor{gray}{rgb}{0.5,0.5,0.5}
\definecolor{mauve}{rgb}{0.58,0,0.82}

\usepackage{multicol}
\usepackage{caption}
\usepackage{subfigure}
\usepackage{microtype}

\usepackage[breaklinks]{hyperref}
\usepackage{url}
\usepackage{lineno}
\begin{document}
\title{TableParser: Automatic Table Parsing with Weak Supervision from Spreadsheets}

\author{
    Susie Xi Rao\textsuperscript{\rm 1}\textsuperscript{\rm 2}\textsuperscript{\rm *},  Johannes Rausch\textsuperscript{\rm 1}\textsuperscript{\rm *}, Peter Egger\textsuperscript{\rm 2}, Ce Zhang\textsuperscript{\rm 1}\\ 
    \textsuperscript{\rm 1} Department of Computer Science, ETH Zurich
    \\
    \textsuperscript{\rm 2} Department of Management, Technology, and Economics, ETH Zurich\\
    \textsuperscript{\rm *} These authors contributed equally to this work. \\
    \{srao,jrausch,pegger,cezhan\}@ethz.ch
}

\maketitle

\begin{abstract}
\begin{quote}
Tables have been an ever-existing structure to store data. There exist now different approaches to store tabular data physically. PDFs, images, spreadsheets, and CSVs are leading examples. Being able to parse table structures and extract content bounded by these structures is of high importance in many applications. In this paper, we devise TableParser, a system capable of parsing tables in both native PDFs and scanned images with high precision. We have conducted extensive experiments to show the efficacy of domain adaptation in developing such a tool. Moreover, we create TableAnnotator and ExcelAnnotator, which constitute a spreadsheet-based weak supervision mechanism and a pipeline to enable table parsing. We share these resources with the research community to facilitate further research in this interesting direction.

\end{quote}
\end{abstract}

\section{Introduction}
Automated processing of electronic documents is a common task in industry and research. 
However, the lack of structures in formats such as native PDF files or scanned documents remains a major obstacle, even for state-of-the-art OCR systems. In practice, extensive engineering and ad-hoc code are required to recover the document structures, e.g., for headings, tables, or nested figures. Sometimes this is required even for text, e.g., in case of PDFs built on the basis of scans, especially, low-quality scans. These structures are hierarchically organized, which many existing systems often fail to recognize. 

With the advance of machine learning (ML) and deep learning (DL) techniques, parsing documents can be done more efficiently than ever. 
As the first end-to-end system for parsing renderings into hierarchical document structures, 
DocParser \citep{rausch21docparser} was recently introduced. It presents a robust way to parse complete document structures from rendered PDFs. Such learning-based systems require large amounts of labeled training data. This problem is alleviated through a novel weak supervision approach that automatically generates training data from structured LaTeX source files in readily available scientific articles. DocParser demonstrates a significant reduction of the labeling complexity through this weak supervision in their experiments. 

As a special document type, tables are one of the most natural ways to organize structured contents. Tabular data are ubiquitous and come in different formats, e.g., CSV (plain and unformatted) and Microsoft Excel (annotated and formatted), depending on the file creation. Many data processing tasks require tables to be represented in a structured format. 
However, structured information is not always available in rendered file formats such as PDF. Especially when PDFs are generated from image scans, such information is missing. Luckily, the existing matrix-type organization of spreadsheets can assist to automatically generate document annotations to PDFs. \textbf{With spreadsheets as weak supervision, this paper proposes a pipeline to provide an automated process of reading tables from PDFs and utilize them as a weak supervision source for DL systems.}

There are three different types of tasks discussed in the literature about table processing in PDFs, namely, table detection, table structure parsing/recognition \citep{zhong2020image}.\footnote{Table  detection is a task to draw the bounding boxes of tables in documents; table structure recognition/parsing refers to the (additional) identification of the structural (row and column layout) information of tables. 
We distinguish between bottom-up and top-down approaches in table structure detection. Bottom-up typically refers to structure detection by recognizing formatting cues such as text, lines, and spacing, while top-down entails table cell detection  (see \cite{kieninger1998t, pivk2007transforming, zhong2020image}).} Table detection is a popular task with a large body of literature, table structure parsing and table recognition were revisited\footnote{Some recent works on Cascade R-CNN \cite{fernandes2022tabledet, prasad2020cascadetabnet} manage to push the frontier of table detection. See \cite{rausch21docparser} for a general review on table detection and \cite{zhong2020image} for a general review on table recognition.} after the pioneering work of \cite{schreiber2017deepdesrt} using state-of-the-art deep neural networks. 
Before DL started to gain success in object detection, table structure parsing was done by bottom-up approaches, using heuristics or ML-based methods like \cite{pinto2003table, farrukh2017interpreting}. See  \cite{pivk2007transforming, wang2004table} for comprehensive reviews on ML methods. The purposes of table structure detection are either layout detection \citep{kieninger1998t} or information retrieval \citep{pivk2007transforming} from tabular structures, usually with the former as a preprocessing step for the latter. 

The DL-based methods in \cite{schreiber2017deepdesrt, qasim2019rethinking} are among the first to apply neural networks designed for object detection to table parsing. Typically, taking pretrained object detection models e.g., Faster RCNN \citep{long2015fully, ren2015faster} on benchmarking datasets like ImageNet \citep{russakovsky2015imagenet}, Pascal VOC \citep{everingham2010pascal}, and Microsoft COCO \citep{lin2015microsoftcoco}, they fine-tune the pretrained models with in-domain images for table detection and table structure parsing (domain adaption and transfer learning). In some best performing frameworks \citep{raja2020table, zheng2021global, jiang2021tabcellnet}, they all jointly optimize the structure detection and entity relations in the structure, as in DocParser. 

However, a key problem in training DL-based systems is the labeling complexity of generating high-quality in-domain annotations. More generally, an essential limiting factor is the lack of large amounts of training data. Efforts have been put into generating datasets to enable tasks with weak supervision. TableBank \citep{li2019tablebank} is built upon a data set of Word and LaTeX files and extracts annotations directly from the sources. They use 4-gram BLEU score to evaluate the cell content alignments. However, the table layout structure is not of particular focus in TableBank. PubTabNet \citep{zhong2020image} enables table detection and table cell content detection. arXivdocs-target and arXivdocs-weak by DocParser \citep{rausch21docparser} enables an end-to-end document parsing system of the hierarchical document structure. 

\begin{figure}[!t]
    \centering
     \includegraphics[width=1\linewidth]{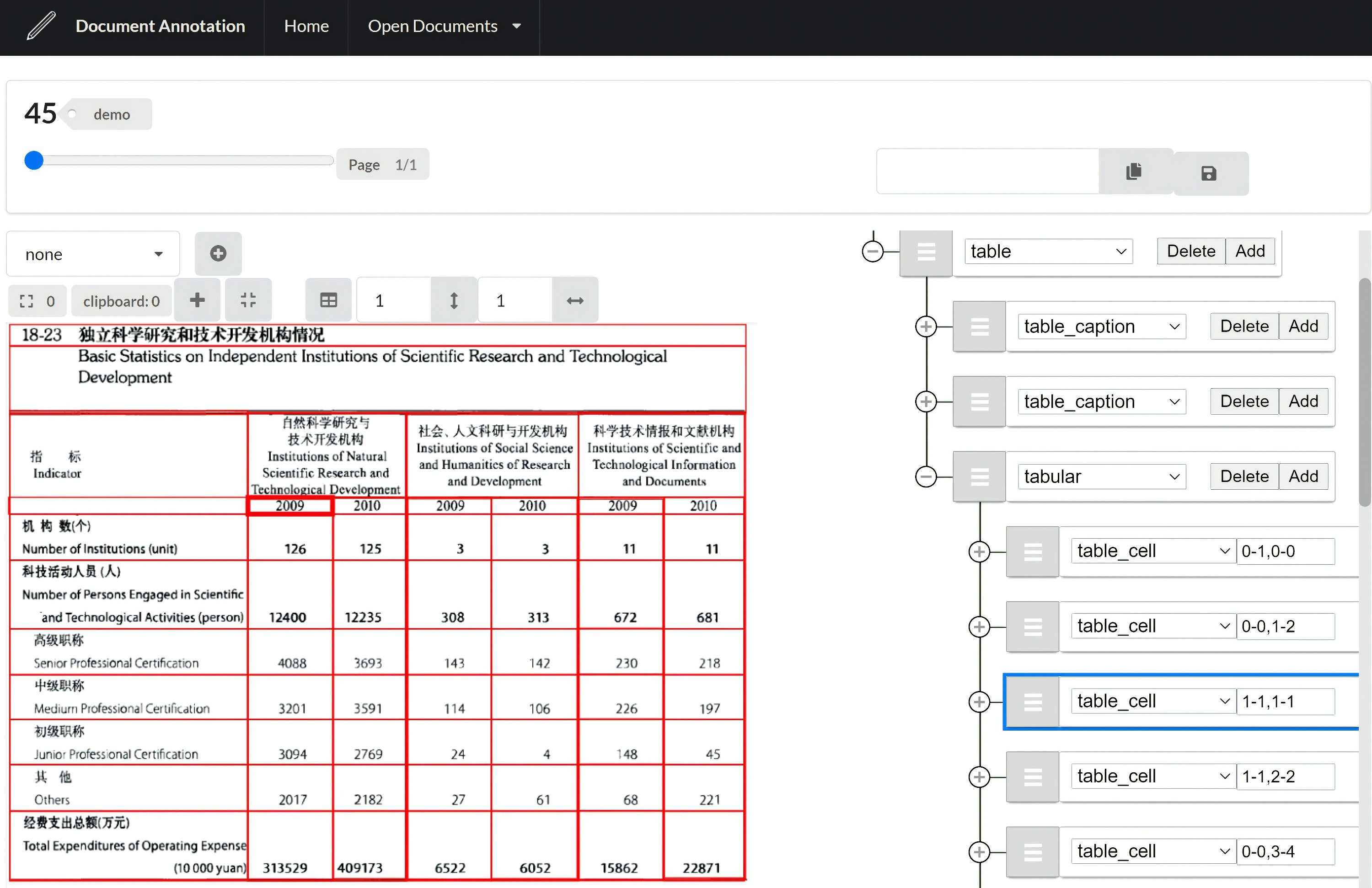}  
    \caption{TableAnnotator.}
    \label{fig:TableAnnotator}
\end{figure}

In this paper, we devise TableParser with inspiration from DocParser, due to its flexibility in processing both tables and more general documents.  
We demonstrate that \textbf{TableParser} is an effective tool for recognizing table structures and content. The application of TableParser to a new target domain requires newly generated training data. 
Depending on the target domain, we specify two TableParsers: \textbf{ModernTableParser} fine-tuned with native PDFs and \textbf{HistoricalTableParser} fine-tuned with scan images. 
TableParser works in conjunction with \textbf{TableAnnotator} (Figure~\ref{fig:TableAnnotator}) which efficiently assists developers in visualizing the output, as well as help users to generate high-quality human annotations.\footnote{For a live demo of table annotations using our annotation tool, refer to the video under \url{https://github.com/DS3Lab/TableParser/blob/main/demo/2021-06-15\%2002-05-58.gif}.} To generate training instances, we develop \textbf{ExcelAnnotator} to interact with spreadsheets and produce annotations for weak supervision. 

\begin{figure*}[!ht]
\centering
\includegraphics[width=1\linewidth]{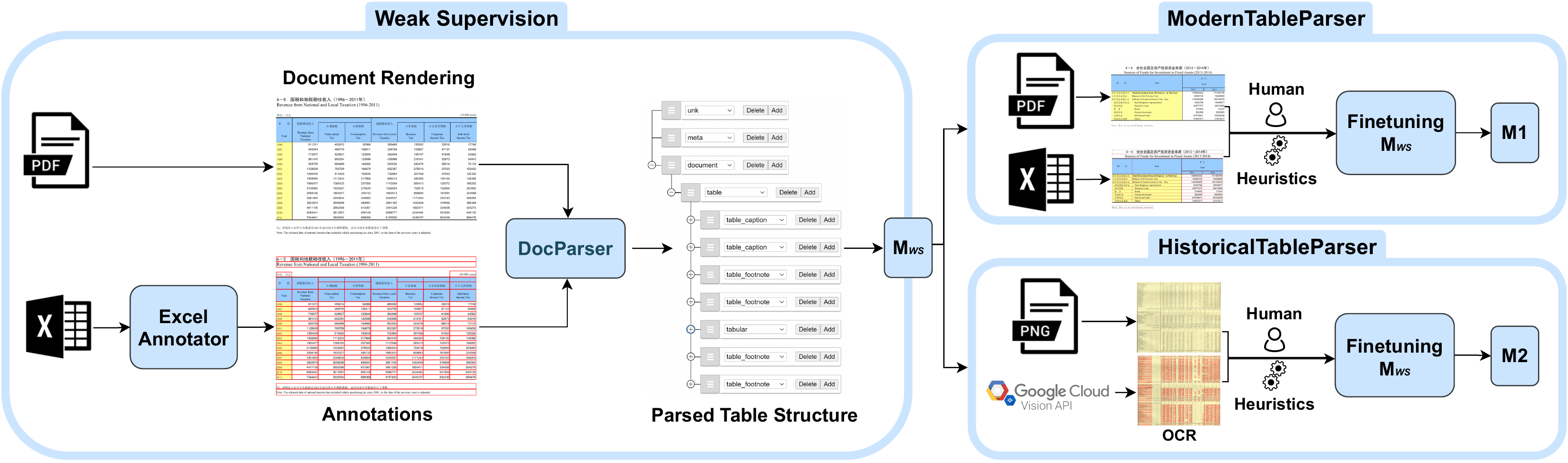} 
\vspace{-1em}
\caption{System design of TableParser: Weak supervision, ExcelAnnotator, ModernTableParser and HistoricalTableParser. $M_{WS}$: the pretrained model with the dataset ZHYearbooks-Excel-WS. M1: for ModernTableParser, fine-tuned on Excel-rendered images; M2: for HistoricalTableParser, fine-tuned on scan images. }
\label{fig:pipeline}
\end{figure*}

With \textbf{ExcelAnnotator}, we have compiled a spreadsheet dataset {ZHYearbooks-Excel}, which is processed via a Python library on Excel (PyWin32\footnote{\url{https://pypi.org/project/pywin32/} (last accessed: Sep. 30, 2021).}) to leverage the structured information stored in the spreadsheets.
TableParser is trained with 16'041 Excel-rendered tables using detectron2 (\cite{he2017maskrcnn, wu2019detectron2}) and fine-tuned with 17 high-quality manual annotations in each domain. We have conducted extensive experiments of domain adaptation. Finally, we evaluate different TableParsers in two domains and make the following observations: 
\begin{enumerate}
    \item In general, domain adaptation works well with fine-tuning the pretrained model ($M_{WS}$ in Figure~\ref{fig:pipeline}) with high-quality in-domain data.
    \item On the test set of 20 tables rendered by Excel, with ModernTableParser we are able to achieve an average precision score (IoU $\geq$ 0.5) of 83.53\% and 73.28\% on table rows and columns, respectively.
    \item We have tested our HistoricalTableParser on scanned tables in both historical (medium-quality, scan-based) and modern tables. Overall, HistoricalTableParser works better than ModernTableParser on tables stored in image scans.  
    \item Interestingly, we find that ModernTableParser built on top of DocParser \citep{rausch21docparser} is very robust in adapting to new domains, such as scanned historical tables. 
\end{enumerate}

We are willing to open source the {ZHYearbook-Excel} dataset, TableAnnotator, TableParser system, and its pipeline to the research communities.\footnote{The source code, data, and/or other artifacts for the complete TableParser pipeline have been made available at \url{https://github.com/DS3Lab/TableParser}.} Moreover, we welcome future contributions to the project to further increase the usability of TableParser in various domains. 

To summarize, our key contributions in this paper are:
\begin{enumerate}
    \item We present \textbf{TableParser} which is a robust tool for parsing modern and historical tables stored in native PDFs or image scans. 
    \item We conduct experiments to show the efficacy of domain adaptation in TableParser. 
    \item We contribute a new pipeline (using \textbf{ExcelAnnotator} as the main component) to automatically generate weakly labeled data for DL-based table parsing.
    \item We contribute \textbf{TableAnnotator} as a graphical interface to assist table structure understanding and manual labeling. 
    \item We open-source the spreadsheet weak supervision dataset and the pipeline of TableParser to encourage further research in this direction.
\end{enumerate}

\begin{figure*}[!ht]
\centering
    \begin{tabular}{cccc}
         \includegraphics[width=0.25\linewidth]{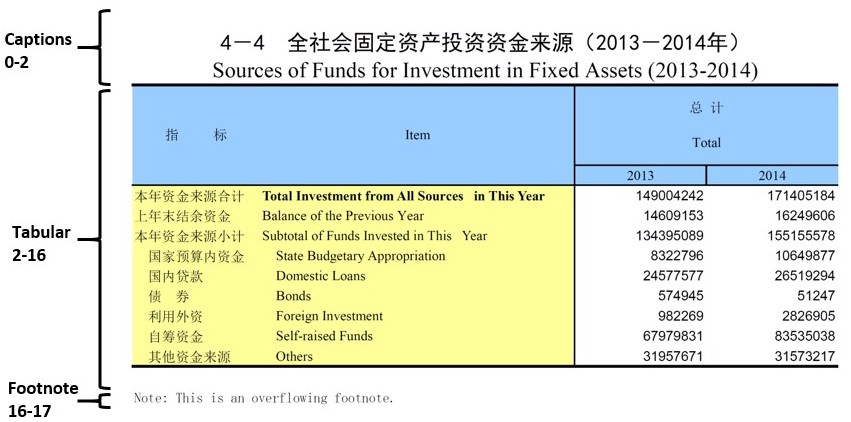} &  \includegraphics[width=0.25\linewidth]{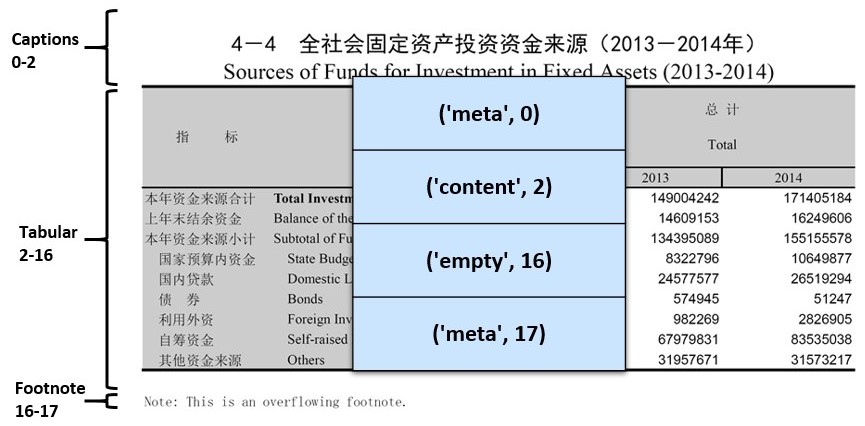}& \includegraphics[width=0.18\linewidth]{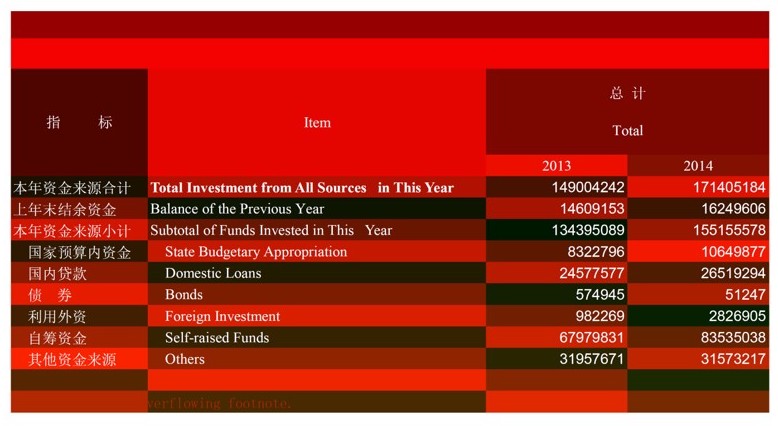}& \includegraphics[width=0.2\linewidth]{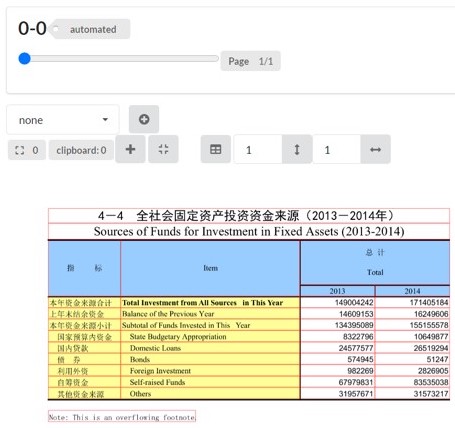} \\
         \begin{tabular}[c]{@{}l@{}}(a) Example worksheet \\from {ZHYearbook-Excel-WS}.
         \end{tabular}
         & 
         { (b) Annotations with DeExcelerator.} 
         & 
         \begin{tabular}[c]{@{}l@{}}(c) Representing bounding \\boxes in Excel.
         \end{tabular}
         &
         \begin{tabular}[c]{@{}l@{}} (d) Visualization of\\ bounding boxes with \\ TableAnnotator.\end{tabular}
        \\
    \end{tabular}
\vspace{-1em}
\caption{A working example in ExcelAnnotator.}
\label{fig:ModernTableParser}
\end{figure*}

\section{TableParser System}\label{sec:sys}

\subsection{Problem Description}
Following the hierarchical document parsing in DocParser, our objective is to generate a hierarchical structure for a table which consists of the entities (\textit{table, tabular, table\_caption, table\_row, table\_column, table\_footnote}) and their relations in the document tree. 

Our ultimate goal of table structure parsing is (1) to establish row/column relationships between the table cells, and (2) post-process the established structure and cell content (e.g., with PDFMiner\footnote{\url{https://pypi.org/project/pdfminer/} (last accessed: Nov. 11, 2021).} or OCR engines) to enable a CSV export function. In this paper, we emphasize (1) and are still in development to enable (2). Our work will enable a user to parse a table stored in a PDF format and obtain (i) the location of a certain cell (specified by its row range and column range) and (ii) the cell content mapped to the cell location. 

\subsection{System Components}
We introduce the main system components in TableParser, incl. TableAnnotator, ExcelAnnotator, ModernTableParser, and HistoricalTableParser. 

\subsubsection{TableAnnotator.}
In Figure \ref{fig:TableAnnotator} we show TableAnnotator, which is mainly composed of two parts: image panel (left) and document tree (right). In the code repository\footnote{TableAnnotator repo: \url{https://anonymous.4open.science/r/doc_annotation-SDU-AAAI22}.}, there is a manual describing its functionalities in details. In a nutshell, annotators can draw bounding boxes on the left panel and create their entities and relationships on the right. In Figure \ref{fig:TableAnnotator}, the highlighted bounding box (the red thick contour on the left) corresponds to the \textit{table\_cell} on the second row and second column, indexed by 1-1, 1-1 (the blue highlight on the right). Note that TableAnnotator is versatile and can be used to annotate not only tables, but also generic documents. The output of the tree is in JSON format as shown in the following code snippet. 


\begin{lstlisting}
  [{"id": 28,
  "category": "table\_cell",
  "properties": "1-1,1-1",
  "row\_range": [1,1],
  "col\_range": [1,1],
  "parent": 9},
  {"id": 29,
  "category": "box",
  "page": 0,
  "bbox": [365,332,299,27],
  "parent": 28}]
\end{lstlisting}

\begin{figure}[!t]
    \begin{subfigure}
    \centering
    \includegraphics[width=0.9\linewidth]{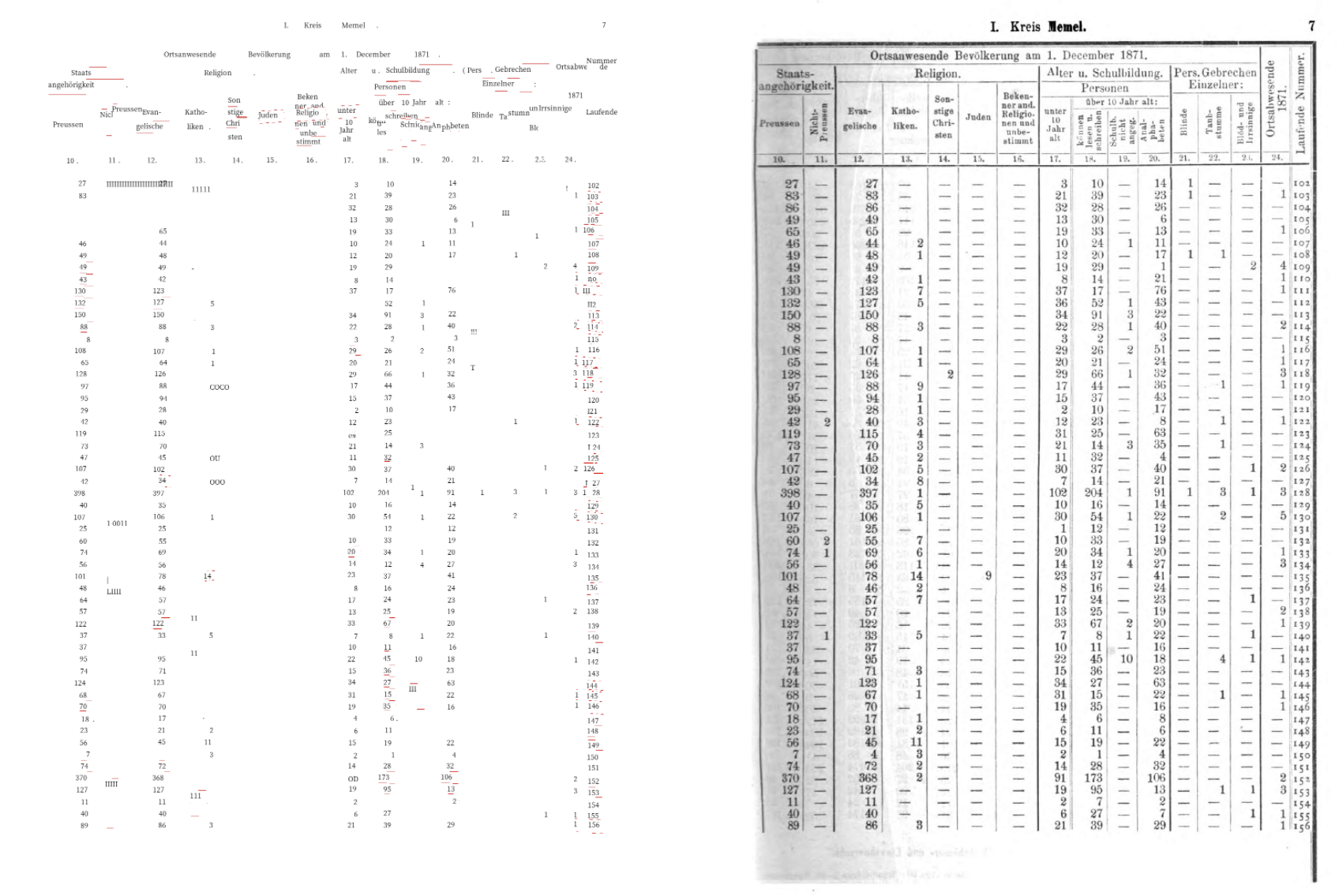}
    \vspace{-1em}
    \caption{Google Vision OCR API output (left) of image (right): bad quality of OCR. }
    \label{fig:bad-google-vision}
    \end{subfigure}
    \begin{subfigure}
    \centering
    \includegraphics[width=0.9\linewidth]{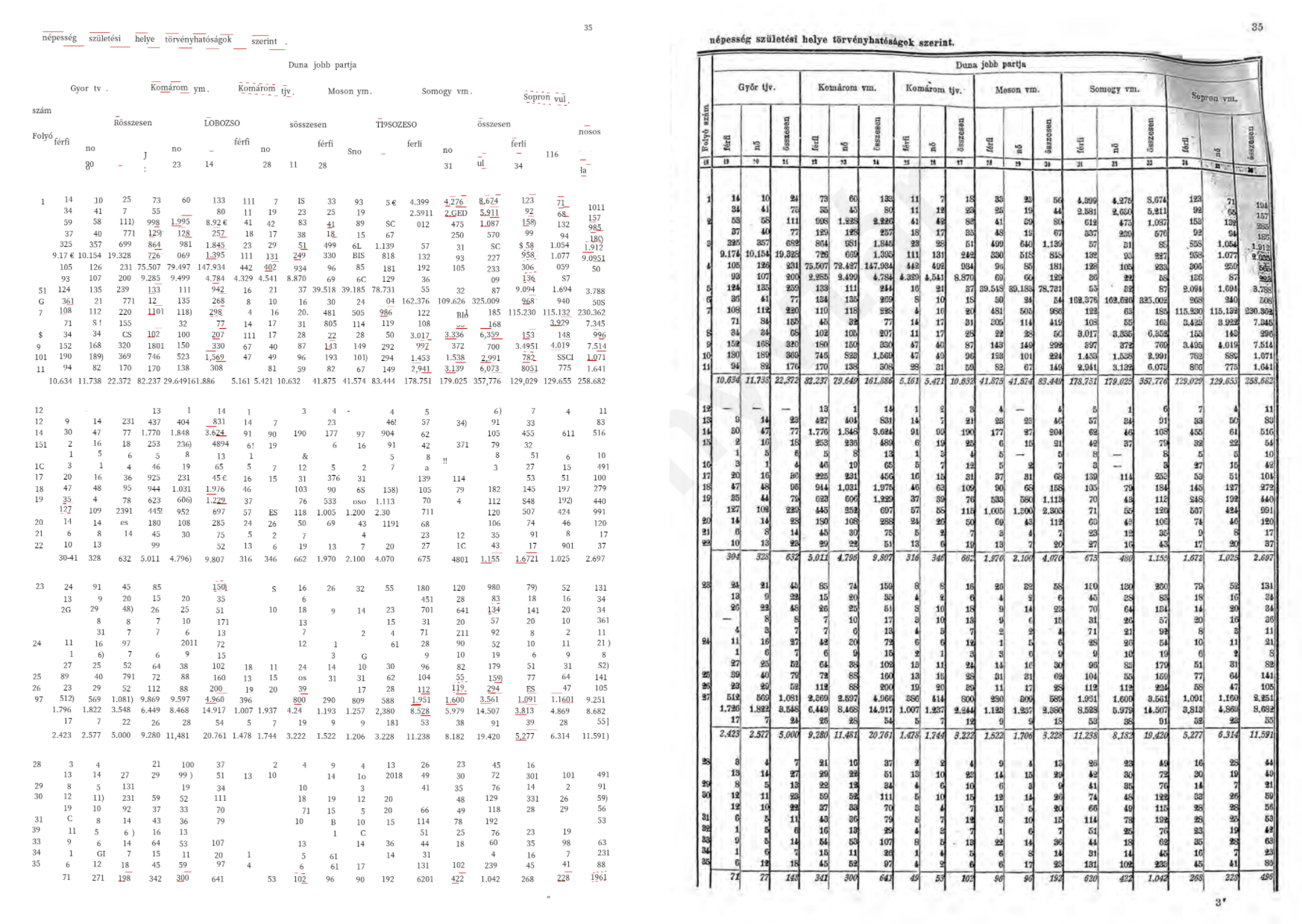}
    \vspace{-1em}
    \caption{Google Vision OCR API output (left) of image (right): good quality of OCR.}
    \label{fig:good-google-vision}
    \end{subfigure}
\end{figure}

\begin{figure*}[!t]
    \centering
    \begin{tabular}{cccc}
       \includegraphics[width=0.2\linewidth]{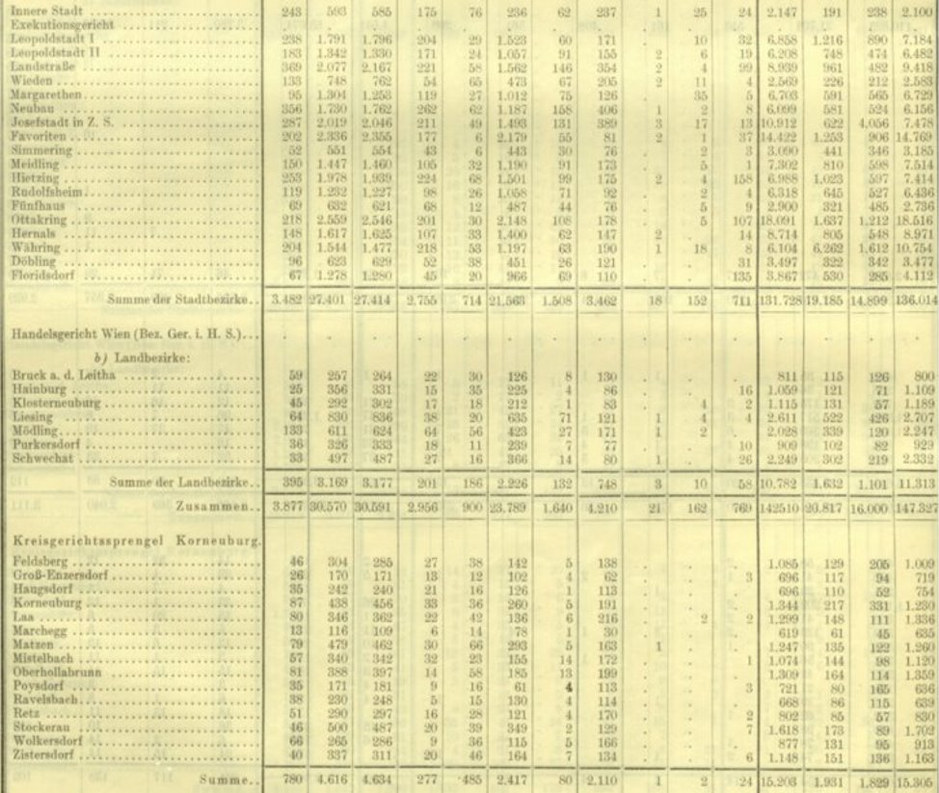}  & \includegraphics[width=0.2\linewidth]{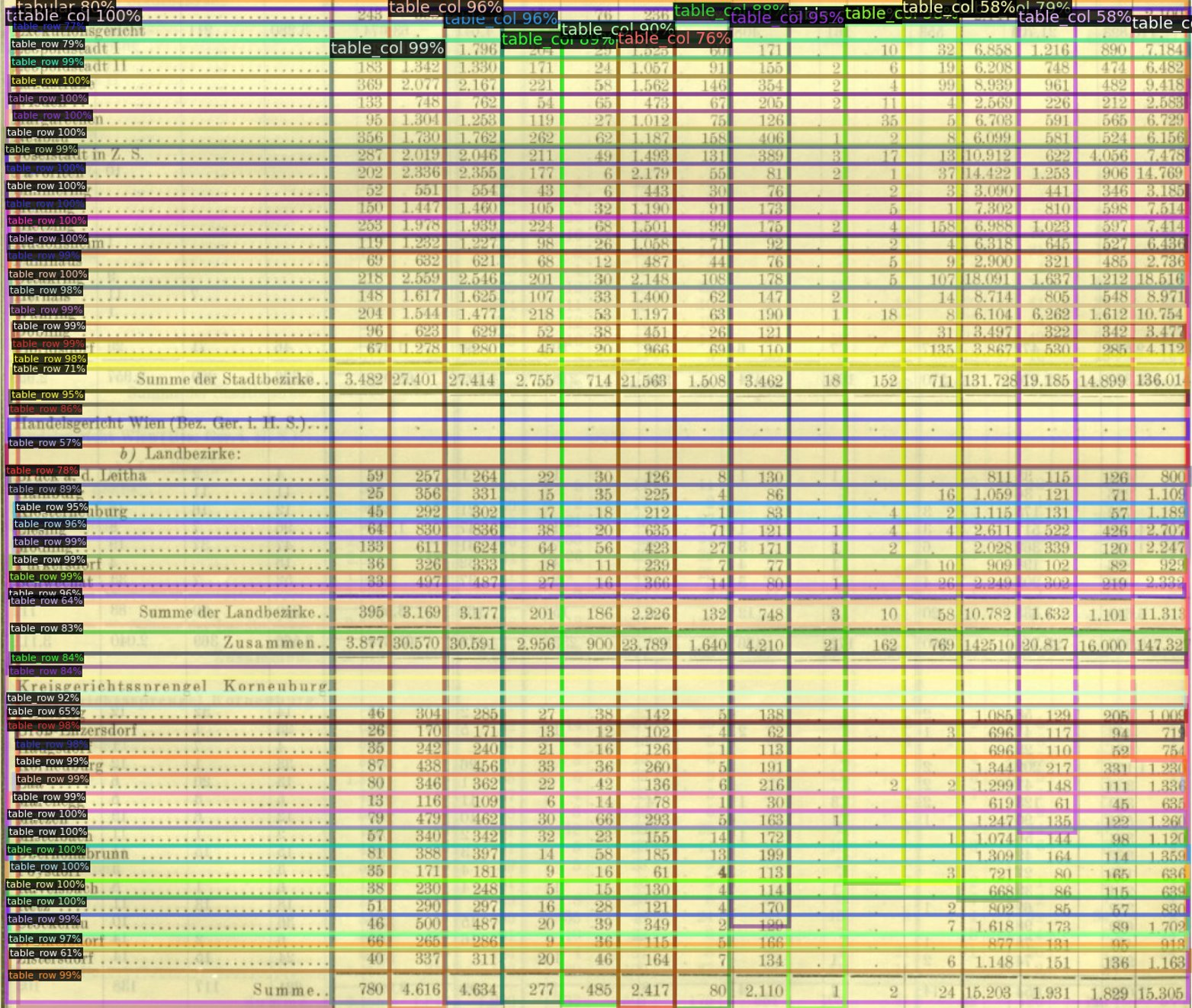}& \includegraphics[width=0.37\linewidth]{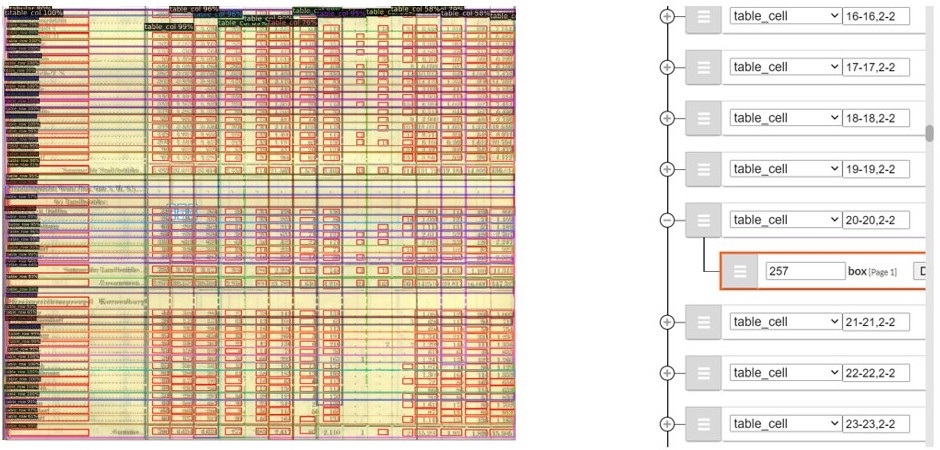} & \includegraphics[width=0.2\linewidth]{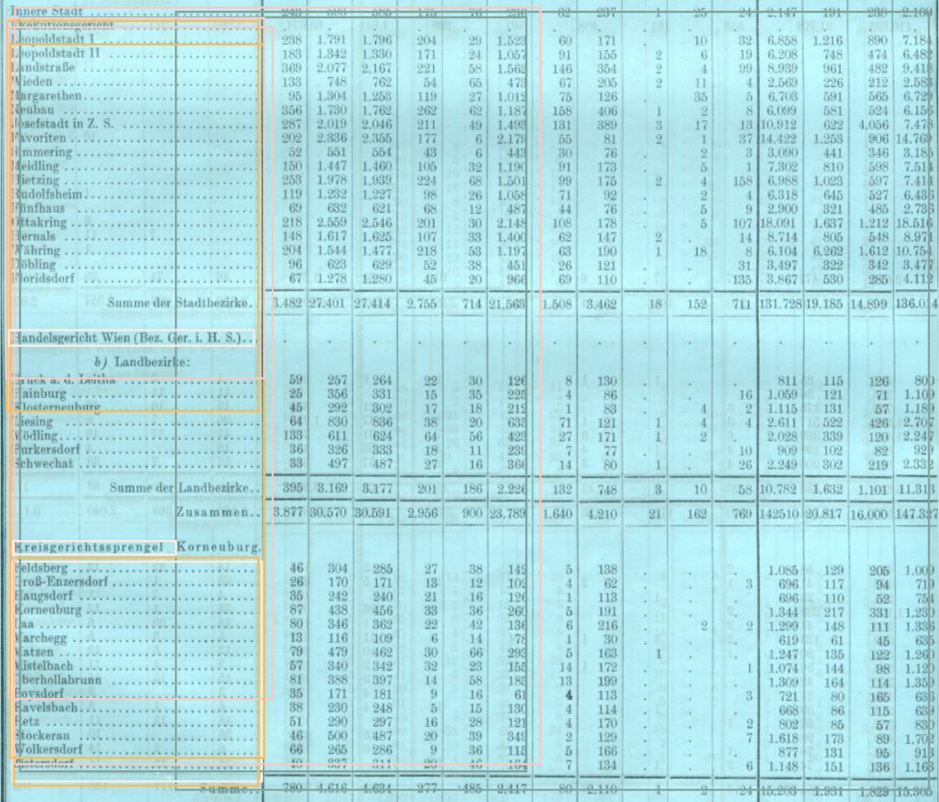} \\
       
       \begin{tabular}[c]{@{}l@{}} (a) An input into\\ HistoricalTableParser.\end{tabular}
       &
       \begin{tabular}[c]{@{}l@{}} (b) Table structure\\ parsing by TableParser.\end{tabular}
       & 
       \begin{tabular}[c]{@{}l@{}} (c) Merging the layout by TableParser \\ and the  OCR bounding boxes.\end{tabular}
       &
       \begin{tabular}[c]{@{}l@{}} (d) Run LayoutParser \\ \citep{shen2021layoutparser}\\ on tables.\end{tabular}
    \end{tabular}
\vspace{-1em}
\caption{A working example in HistoricalTableParser.}
\label{fig:HistoricalTableParser}
\end{figure*}

\subsubsection{ModernTableParser.}

We train ModernTableParser using the data generated by weak supervision signals from Excel sheets and fine-tuned by high-quality manual annotations in this domain. In Figure \ref{fig:pipeline}, we show the system design following the underlying components of DocParser.\footnote{The model structure of DocParser is sketched in Figure 11 of the DocParser paper \citep{rausch21docparser}, see \url{https://arxiv.org/pdf/1911.01702.pdf}. The model structure (Mask RCNN) can also be found \href{https://github.com/DS3Lab/TableParser/blob/main/figures/mask-rcnn.drawio.pdf}{here}.} We denote the model that produces ModernTableParser as \textbf{M1}. 

\paragraph{Weak Supervision with ExcelAnnotator.} Now we present the crucial steps in generating weak supervision (the model $M_{WS}$ in Figure \ref{fig:pipeline}) for TableParser. These steps are mainly conducted by ExcelAnnotator in Figure \ref{fig:pipeline} (left). Take a worksheet-like Figure \ref{fig:ModernTableParser} (a) from our {ZHYearbook-Excel-WS} dataset (cf. Section \ref{sec:data}), where we see caption, tabular, and footnote areas. We subsequently use DeExcelerator \citep{eberius2013deexcelerator} 
to extract relations from the spreadsheets.
We utilize DeExcelerator to categorize the content, such that we can differentiate among table captions, table footnotes and tabular data and create a correct auxiliary file to each PDF containing the structural information of the represented table(s). 
Illustrated in Figure \ref{fig:ModernTableParser} (b), in this case we annotate the table caption and footnote as `meta', and mark the range of content with `content' and `empty'. We use PyWin32 in Python to interact with Excel, so that intermediate representations like Figure \ref{fig:ModernTableParser} (c) can be created to retrieve entity locations in the PDF rendering. Concretely, we mark neighboring cells with distinct colors, remove all borders, and set the font color to white. To summarize, ExcelAnnotator detects spreadsheet metadata and cell types, as well as retrieves entity locations via intermediate representations. Finally, we are able to load the annotations into TableAnnotator to inspect the quality of weak supervision (Figure \ref{fig:ModernTableParser} (d)). 

\subsubsection{HistoricalTableParser.}
We use the OCR engine from Google Vision API to recognize the text bounding boxes. Then we convert bounding boxes into the input format TableParser requires. Now we are able to manually adjust the bounding boxes in TableAnnotator to produce high-quality annotations. Note that the quality of OCR highly depends on the table layout (Figures \ref{fig:bad-google-vision} vs. \ref{fig:good-google-vision}), we often need to adjust the locations of bounding boxes and redraw the bounding boxes of individual cells. 

In Figure \ref{fig:pipeline} (lower right), we show the system design by adding an OCR component and a fine-tuning component for domain adaptation. We denote the model that produces HistoricalTableParser as \textbf{M2}. 
Take Figure \ref{fig:HistoricalTableParser} (a) as input, TableParser can produce a parsed layout-like Figure \ref{fig:HistoricalTableParser} (b) which can be combined with the OCR bounding boxes in the subsequent steps and export as a CSV file (Figure \ref{fig:HistoricalTableParser} (c)).\footnote{c.f. The performance of LayoutParser is quite poor on the tabular data in Figure \ref{fig:HistoricalTableParser} (d) using the best model from its model zoo (PubLayNet/faster\_rcnn\_R\_50\_FPN\_3x). Input and annotated figures of original size can be found under \url{https://github.com/DS3Lab/TableParser/tree/main/figures}.} 

For domain adaptation, we assume that an out-of-domain model performs worse than an in-domain model in one domain. Namely, we would expect ModernTableParser to work better on Excel-rendered PDFs or tables created similarly; on the contrary, we would expect HistoricalTableParser to perform better on older table scans.

\section{Datasets}\label{sec:data}
We have compiled various datasets to train, fine-tune, test, and evaluate TableParser.

\textbf{ZHYearbooks-Excel.}
We create three datasets from this source: {ZHYearbooks-Excel-WS},  {ZHYearbooks-Excel-FT}, and {ZHYearbooks-Excel-Test}, with 16'041, 17, and 20 tables in each set. 
On average, it takes 3 minutes 30 seconds for an annotator to produce high-quality annotations of a table. The manual annotations are done with automatically generated bounding boxes and document tree as aid.

\textbf{ZHYearbooks-OCR.}
We create the dataset {ZHYearbook-OCR-Test}, with 20 tables. On average, it takes 2 minutes and 45 seconds to annotate a table with the similar annotation aids mentioned above.

\textbf{EUYearbooks-OCR.}
We create two datasets from this source: {EUYearbook-OCR-FT} and {EUYearbook-OCR-Test}, with 17 and 10 tables, respectively. Note that these datasets contain various languages like Hungarian and German, with various formats depending on the language. On average, it takes 8 minutes and 15 seconds to annotate a table with the similar annotation aids mentioned above.

\textbf{Miscellaneous historical yearbooks.}
We ran ModernTableParser and HistoricalTableParser on Chinese and Korean historical yearbooks and inspect their outputs qualitatively (see Section \ref{sec:quali}).  

\textbf{Human labeling efforts.}
We observe a large variance in labeling intensity across the datasets. The EUYearbooks-OCR datasets require more corrections per table compared to the datasets of modern tables. Moreover, they also require more iterations of human annotations with heuristics as aid.

\section{Computational Setup}
\subsection{Mask R-CNN} \label{sec:maskrcnn}
In line with DocParser, we use the same model but with an updated backend implementation. Namely, we utilize Detectron2 to apply an updated version of Mask R-CNN \citep{he2016deep}. For technical details of Mask R-CNN, we refer to DocParser \citep{rausch21docparser}.

\subsubsection{Training Procedure: Weak Supervision + Fine-Tuning.}
All neural models are initialized with weights trained on the MS COCO dataset. We first pretrain on the weak supervision data {ZHYearbook-Excel-WS} for 540k iterations, then fine-tune on our target datasets {ZHYearbook-Excel-FT} and {EUYearbook-OCR-FT} for M1 and M2, respectively. 
We then fine-tune each model across three phrases for a total of 30k iterations. This is split into 22k, 4k, 4k iterations, respectively. The performance is measured every 500 iterations via the IoU with a threshold of 0.5. We train all models in a multi-GPU setting, using 8 GPUs with a vRAM of 12 GB. Each GPU was fed with one image  per  training  iteration.  Accordingly,  the  batch  size per training iteration is set to 8. Furthermore, we use stochastic gradient descent with a learning rate of 0.005 and learning momentum of 0.9.

\subsubsection{Parameter Settings.}

During training, we sampled randomly 100 entities from the ground truth per document image (i.e., up  to  100  entities,  as  some  document  images  might  have less).  In  Mask  R-CNN,  the  maximum  number  of  entity predictions per image is set to 100. During prediction, we only  keep  entities  with  a  confidence  score of 0.5 or higher. 
\begin{table*}[!t]
\centering
\caption{Fine-tuning results of M1 and M2.  
{M1: for ModernTableParser, fine-tuned on Excel-rendered images; M2: for HistoricalTableParser, fine-tuned on scan images; FT: fine-tune.}}
\vspace{-1em}
\label{tab:ft-results}
\resizebox{0.65\linewidth}{!}{
\begin{tabular}{c|c|c|c|c|c|c|c}
\toprule
\multicolumn{4}{c|}{\textbf{ZHYearbook-Excel-FT}}                              & \multicolumn{4}{c}{\textbf{EUYearbook-OCR-FT}}                                \\ 
\midrule
\multirow{2}{*}{\textbf{Category}} & \multirow{2}{*}{\textbf{\# instances}} & \multicolumn{2}{c|}{\underline{\textbf{Average Precision}}} & \multirow{2}{*}{\textbf{Category}} & \multirow{2}{*}{\textbf{\# instances}} & \multicolumn{2}{c}{\underline{\textbf{Average Precision}}} \\   
                          &                              & \textbf{M1} (FT)      & \textbf{M2} (Test)   &                           &                              & \textbf{M1} (Test)   & \textbf{M2} (FT)     \\ \midrule
(1) & (2) & (3) & (4) & (5) & (6) & (7) & (8) \\ \midrule
\textit{table}                     & 17                           & 90.973   & 38.034      & \textit{table}                     & 17                           & 67.467      & 93.011   \\
\textit{tabular}                   & 17                           & 100.000      & 57.897      & \textit{tabular}                   & 17                           & 76.423      & 100.000      \\
\textit{table\_column}                & 134                          & 96.730    & 15.253      & \textit{table\_column}                & 260                          & 24.930       & 81.376   \\
\textit{table\_row}                & 548                          & 79.228   & 39.485      & \textit{table\_row}                & 1180                         & 19.256      & 60.899  \\ \bottomrule
\end{tabular}
}
\end{table*}

\begin{table*}[!t]
\centering
\caption{Test results of M1 and M2 on various data sets.}
\vspace{-1em}
\label{tab:test-results}
\resizebox{0.9\textwidth}{!}{
\begin{tabular}{c|c|c|c|c|c|c|c|c|c|c|c}
\toprule
\multicolumn{4}{c|}{\textbf{ZHYearbook-Excel-Test}}      & \multicolumn{4}{c|}{\textbf{ZHYearbook-OCR-Test}}      & \multicolumn{4}{c}{\textbf{EUYearbook-OCR-Test}}      \\ \midrule
\multirow{2}{*}{\textbf{Category}}     & \multirow{2}{*}{\textbf{\# instances}}   & \multicolumn{2}{l|}{\underline{\textbf{Average Precision}}} & \multirow{2}{*}{\textbf{Category} }   & \multirow{2}{*}{\textbf{\# instances}}  & \multicolumn{2}{l|}{\underline{\textbf{Average Precision}}} & \multirow{2}{*}{\textbf{Category}}    & \multirow{2}{*}{\textbf{\# instances}}  & \multicolumn{2}{l}{\underline{\textbf{Average Precision}}} \\
             &               & \textbf{M1}         & \textbf{M2}        &             &              & \textbf{M1}         & \textbf{M2}        &             &              & \textbf{M1}         & \textbf{M2}        \\ \midrule
             (1) & (2) & (3) & (4) & (5) & (6) & (7) & (8) & (9) & (10) & (11) & (12) \\ \midrule
\textit{table}        & 20            & 85.407     & 32.821    & \textit{table}       & 10           & 56.942     & 53.356    & table       & 10           & 57.151     & 81.907    \\
\textit{tabular}      & 21            & 80.193     & 43.801    & \textit{tabular}     & 10           & 64.175     & 52.563    & \textit{tabular}     & 10           & 85.956     & 91.429    \\
\textit{table\_column}   & 176           & 73.277     & 14.927     & \textit{table\_column}  & 74           & 43.094     & 21.997    & \textit{table\_column}  & 136          & 36.616     & 40.509    \\
\textit{table\_row}   & 513           & 83.528     & 48.912    & \textit{table\_row}  & 226          & 50.055     & 36.619    & \textit{table\_row}  & 665          & 25.645     & 40.229   \\ \bottomrule
\end{tabular}
}
\end{table*}

\section{Results and Discussion}
Here, we evaluate the performance of TableParser in two domains quantitatively and qualitatively. 

\subsection{Quantitative assessment}

\paragraph{Metric.} We first introduce the evaluation metric for the object detection/classification tasks. The metric we report is Average Precision (AP), which corresponds to an Intersection over Union rate of IoU=.50:.05:.95.\footnote{We refer readers to \url{https://cocodataset.org/\#detection-eval} for more details on the evaluation metrics (last accessed: Nov. 1, 2021).} IoU ranges from 0 to 1 and specifies the amount of overlap between the predicted and ground truth bounding box. It is a common metric used when calculating AP.

\paragraph{Performances in various domains.} As we discussed in Section~\ref{sec:sys}, we have developed ModernTableParser to parse tables with input images rendered by Excel (M1). Then, to work with historical tables in scans, we adapt the pretrained TableParser by fine-tuning it on scanned documents (M2). Now, we present the performances of M1 and M2 in two different domains in the following aspects:
\begin{enumerate}
    \item \textbf{(P1)} the performances on fine-tuning sets on M1 and M2 in Table~\ref{tab:ft-results};
    \item \textbf{(P2)} the performances on fine-tuning sets as test sets on M1 and M2 in Table~\ref{tab:ft-results};\footnote{This means we evaluate the performance of M1 on the fine-tuned set for M2 (as a test set for M1) and vice versa.}
    \item \textbf{(P3)} the performances on three test sets from two domains on M1 and M2 in Table~\ref{tab:test-results}. 
\end{enumerate}

\textbf{(P1) \& (P2).}
We want to study the impact of fine-tuning of a pretrained model (using a large body of tables generated by weak supervision signals). The instances used to fine-tune must be high-quality in-domain data. Concretely, we create in-domain annotations for modern tables (rendered by Excel) and historical tables (from scans) with high human efforts assisted by automatic preprocessing: {ZHYearbook-Excel-FT} and {EUYearbook-OCR-FT}, each with 17 tables. Note that the latter has much denser rows and columns than the former (see the tables in Figures~\ref{fig:ModernTableParser} (a) vs.~\ref{fig:HistoricalTableParser} (a) for an illustration). It is apparent from Table~\ref{tab:ft-results} that the AP performance of models on the fine-tuning sets is highly optimized (columns (3) and (8) in Table~\ref{tab:ft-results}), and it should be better than using those datasets as test sets. This means, if we run M1 (fine-tuned by modern tables) on {EUYearbook-OCR-FT} (column 7), its performance is worse than fine-tuning; and if we run M2 (fine-tuned by historical tables) on {ZHYearbook-Excel-FT} (column 4), it performs worse than fine-tuning. Interestingly, if we compare the performance of M2 on modern tables (column (4)) with the performance of M1 on historical tables (column (7)), we clearly see that the latter has a better performance in all other categories than the class of \textit{table\_row}. This can be explained by the fact that the model trained on modern tables is robust in annotating historical tables, at least on the column level. We see this in Figures \ref{fig:M1-vs-M2_KO} and \ref{fig:M1-vs-M2_ZH}, where ModernTableParser clearly performs better. \textbf{However, the algorithm has problems in delineating narrow and less clearly separated rows.} This could be due to the setting of the maximum number of entities being 100 when predicting per table (Section \ref{sec:maskrcnn}).

\textbf{(P3).} In Table~\ref{tab:test-results}, we show the performances of three test sets from two domains (Excel-rendered PDFs and historical scans), namely, {ZHYearbook-Excel-Test}, {ZHYearbook-OCR-Test}, and {EUYearbook-OCR-Test}. We see that M2 which is fine-tuned by historical scans performs worse than M1 on both {ZHYearbook-Excel-Test} and {ZHYearbook-OCR-Test}. Vice versa, M1 that is fine-tuned by Excel-rendered PDFs performs worse than M2 on {EUYearbook-OCR-Test}. This suggests that domain adaptation by fine-tuning the pretrained TableParser with in-domain high-quality data works well.  

Additionally, if we compare the $\Delta AP|(M1-M2)|$ under each test set (e.g., the differences of columns (3) and (4), of (7) and (8), of (11) and (12)), the $\Delta AP$ on {*-OCR-Test} in all categories is smaller than {ZHYearbook-Excel-Test}, with M1 already achieving medium-high performance on the test set. Although M1 is not fine-tuned by in-domain historical images, ModernTableParser is still able to parse historical scans with moderate performance. This suggests that TableParser trained on modern table structures can be used to parse the layout of tabular historical scans. Because the cost is often too high in generating a large amount of training data of historical scans (see Section \ref{sec:data} for the discussion of labeling efforts), our approach shows a promising direction in first developing TableParser that works well for modern tables, and then adapting TableParser to the historical domain by fine-tuning on only a few manually annotated historical scans of good quality.

\subsection{Qualitative Assessment} \label{sec:quali}

\begin{figure}[!t]
    \centering
    \begin{tabular}{cc}
    \includegraphics[width=0.5\linewidth]{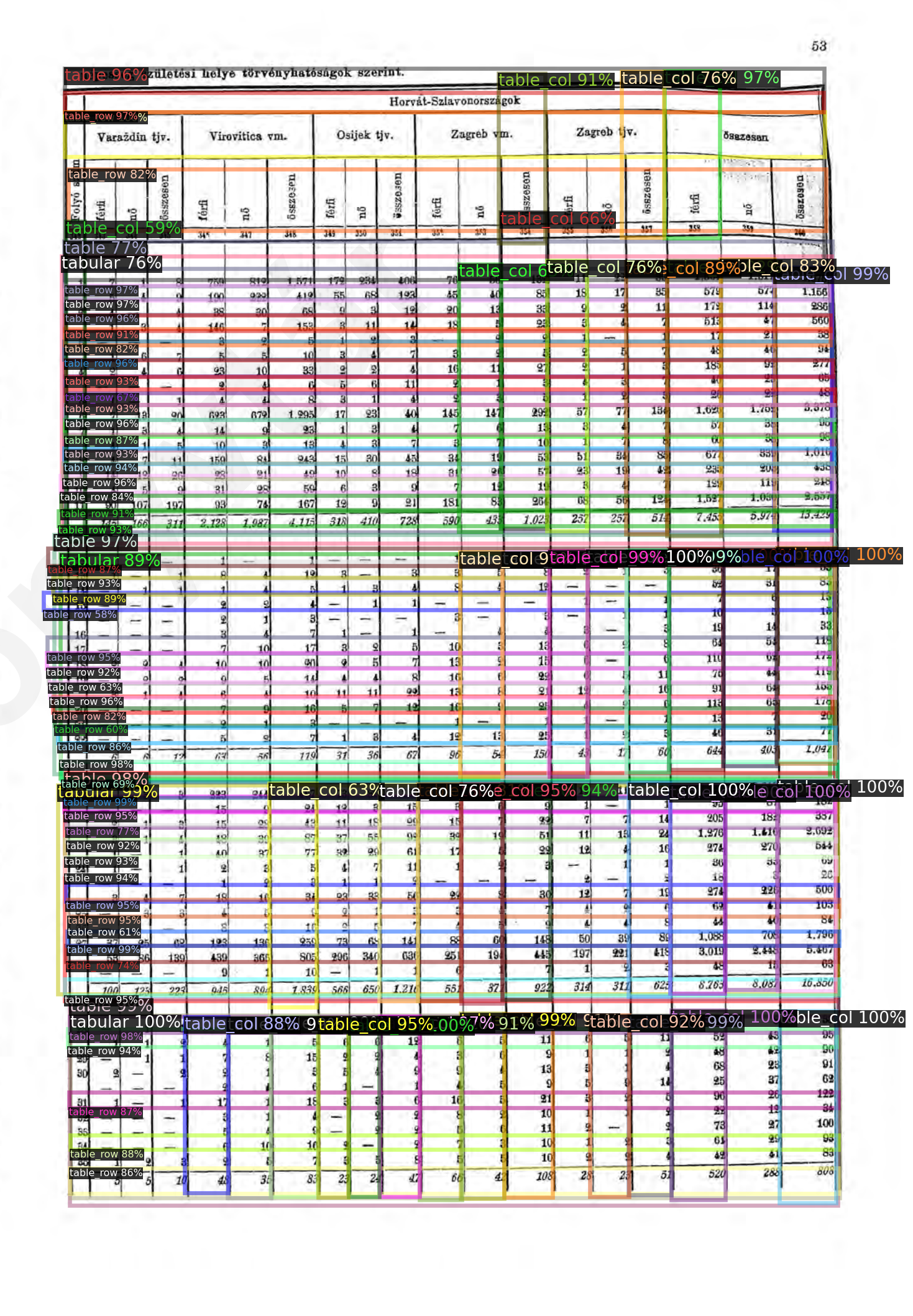}

& 
    \includegraphics[width=0.5\linewidth]{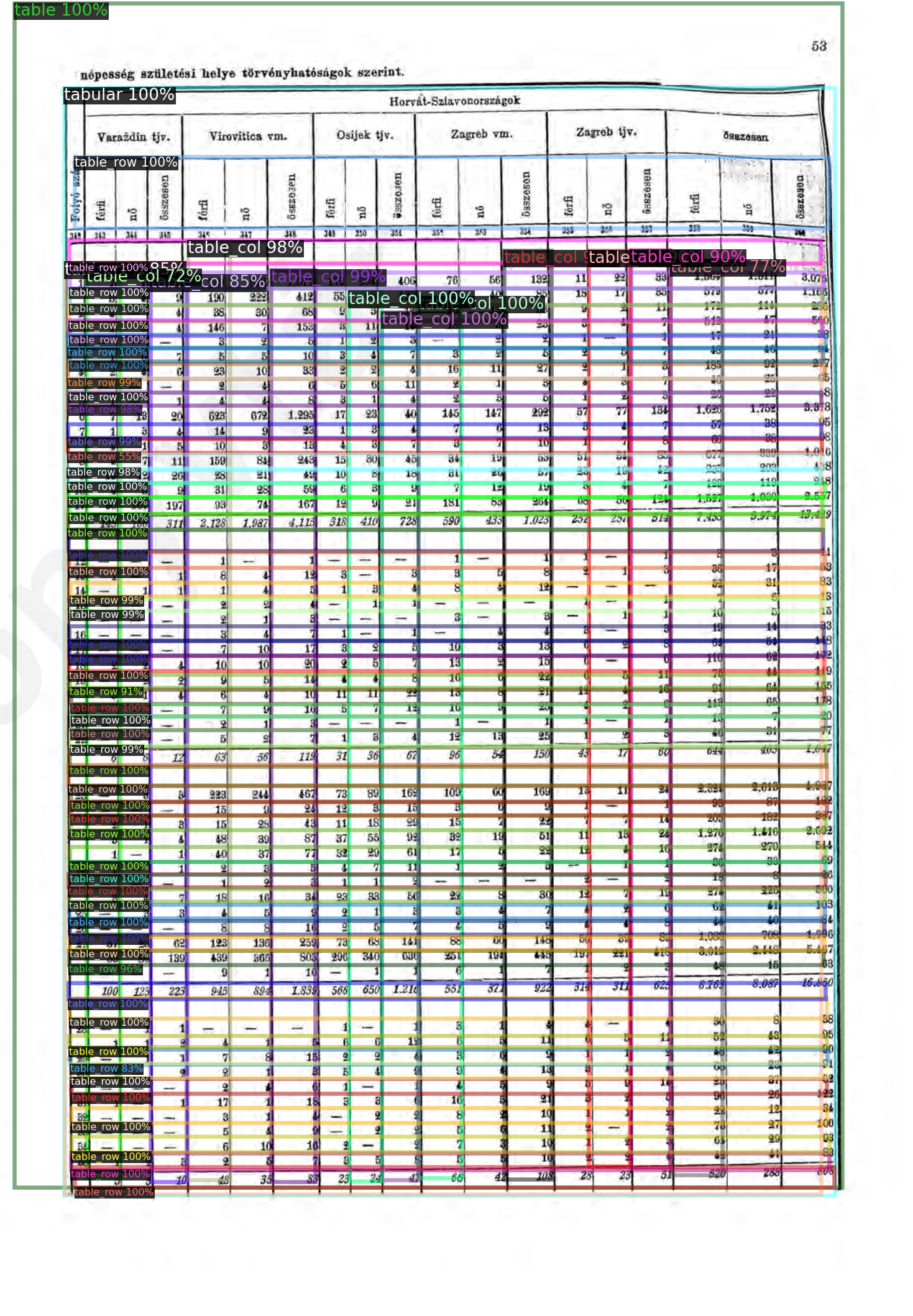}
    \end{tabular}
\vspace{-1em}
\caption{A Hungarian table parsed by ModernTableParser (left) and HistoricalTableParser (right).}
\label{fig:M1-vs-M2_18-0}
\end{figure}

\begin{figure}[!t]
    \centering
    \begin{tabular}{cc}
    \includegraphics[width=0.48\linewidth]{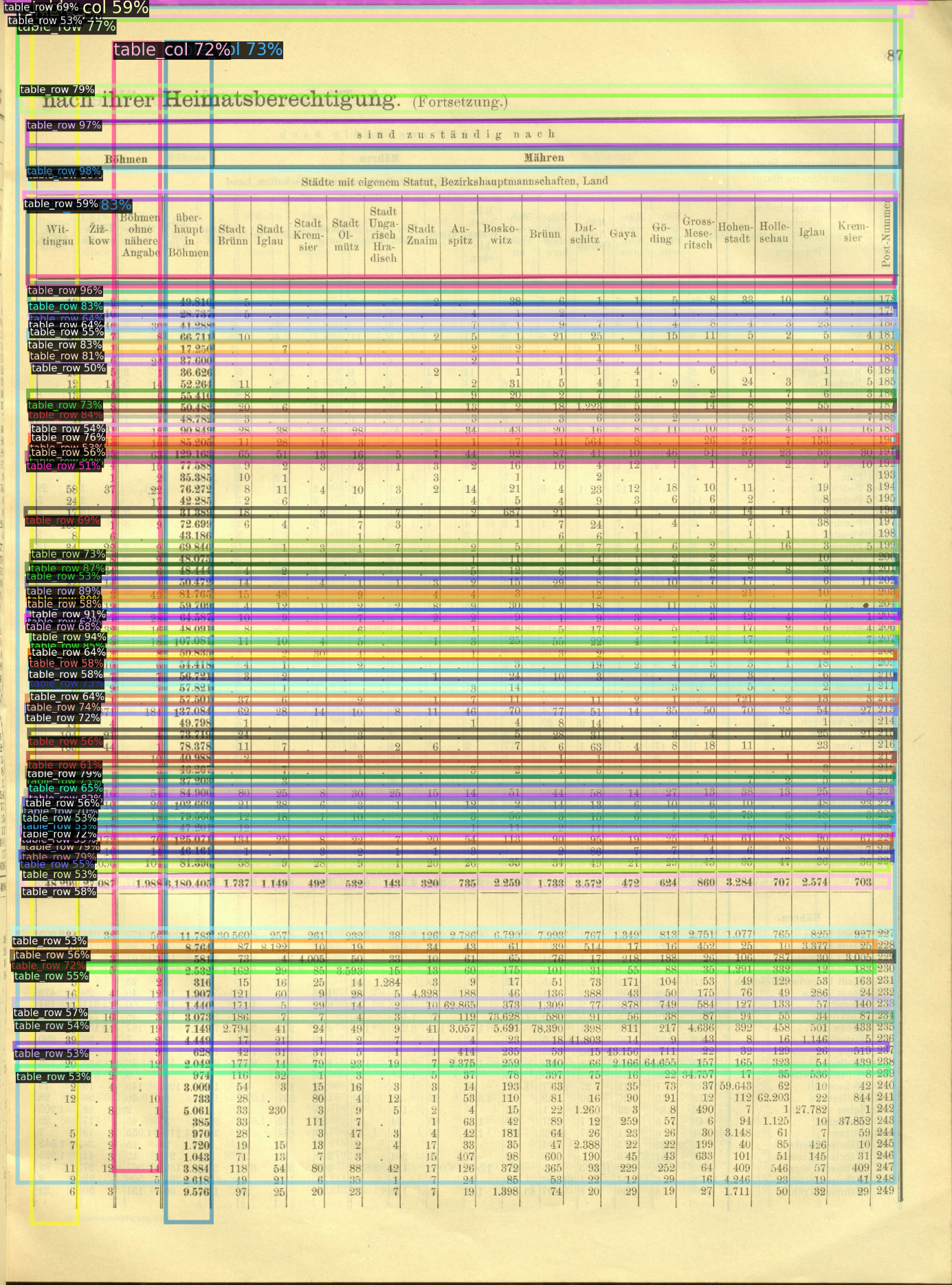}
    & 
    \includegraphics[width=0.48\linewidth]{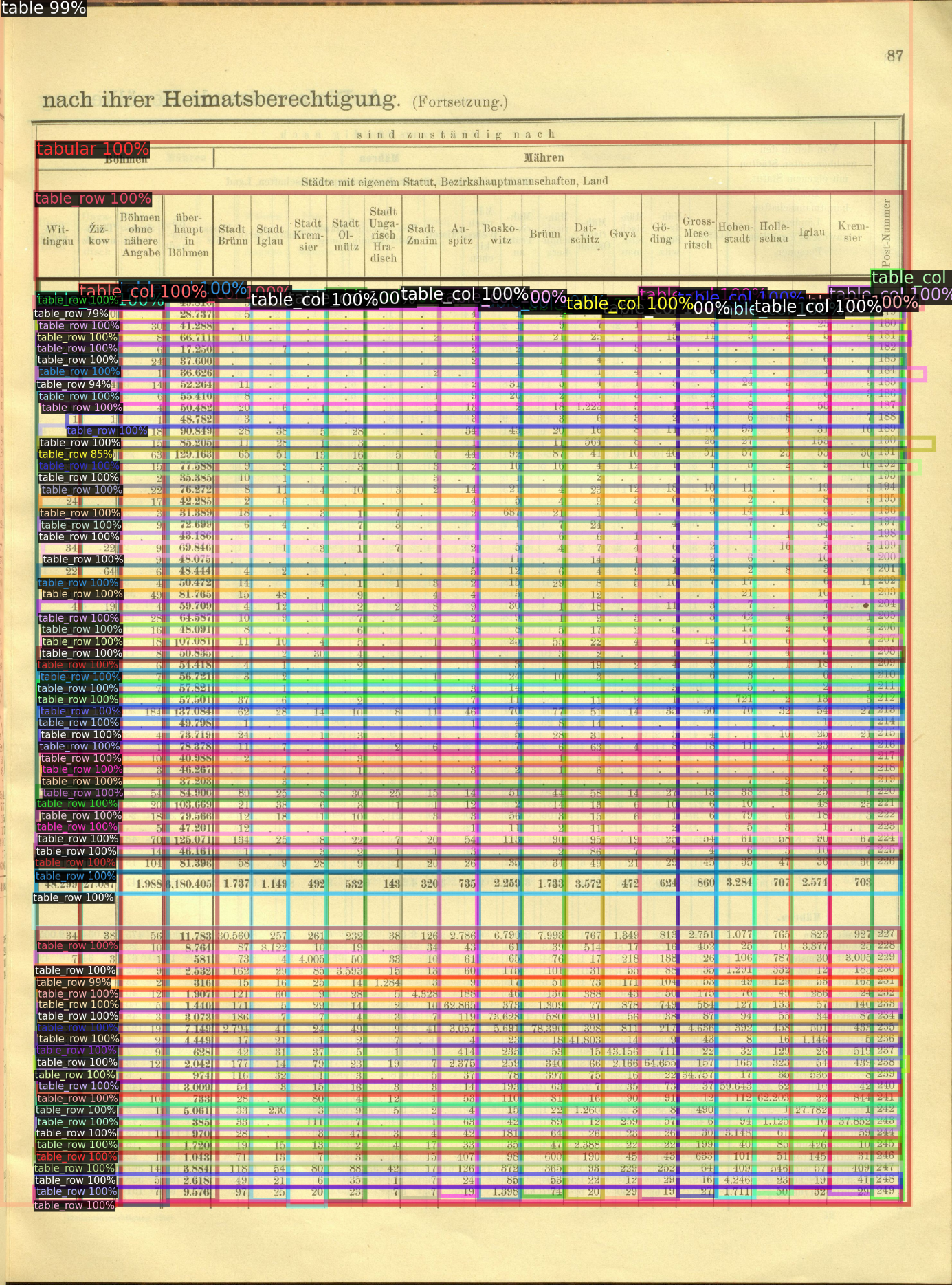}
    \end{tabular}
    \vspace{-1em}
    \caption{A German table parsed by ModernTableParser (left) and HistoricalTableParser (right).}
    \label{fig:M1-vs-M2_27-0}
\end{figure}

\begin{figure}[ht]
 \centering
 \begin{tabular}{cc}
     \centering
    \includegraphics[width=0.48\linewidth]{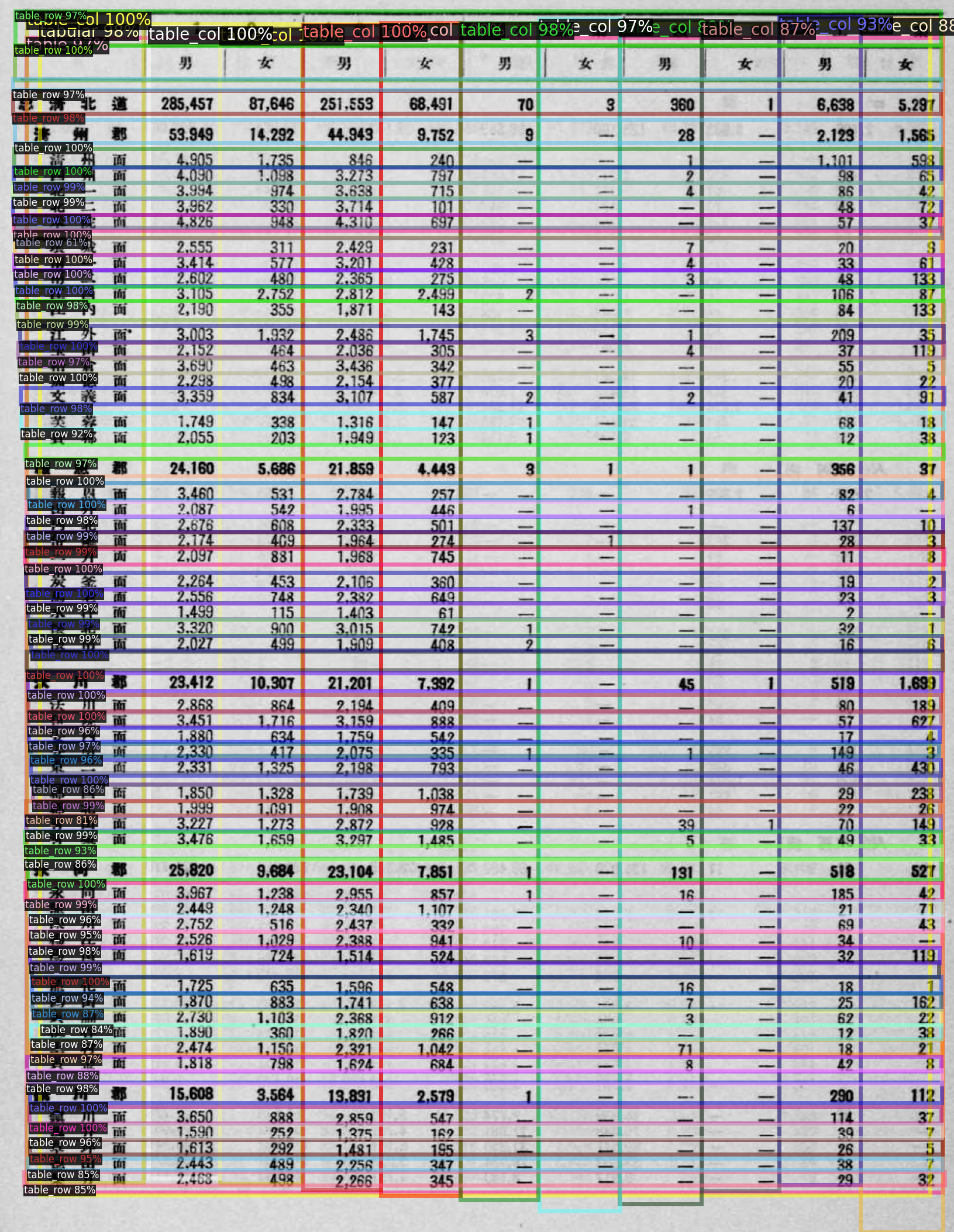}

    &
        \centering
    \includegraphics[width=0.48\linewidth]{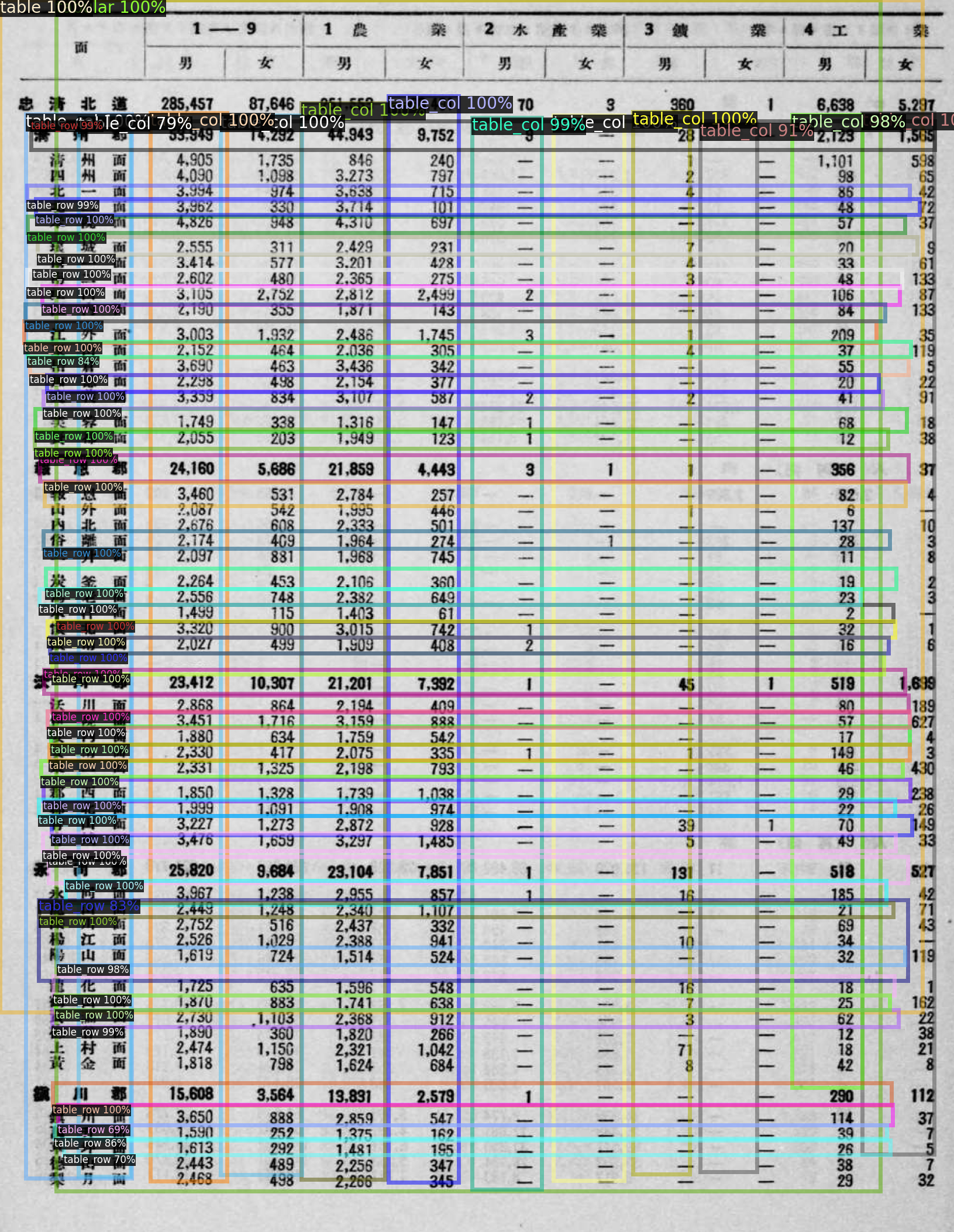}
 \end{tabular}
 \vspace{-1em}
    \caption{A Korean table parsed by ModernTableParser (left) and HistoricalTableParser (right).}
    \label{fig:M1-vs-M2_KO}
\end{figure}

\begin{figure}[ht]
    \centering
    \begin{tabular}{cc}
    \centering
    \includegraphics[width=0.24\linewidth]{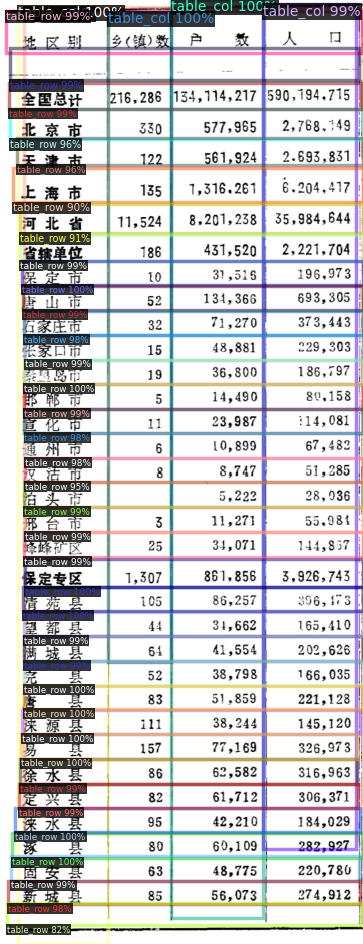}
    &
    \centering
    \includegraphics[width=0.24\linewidth]{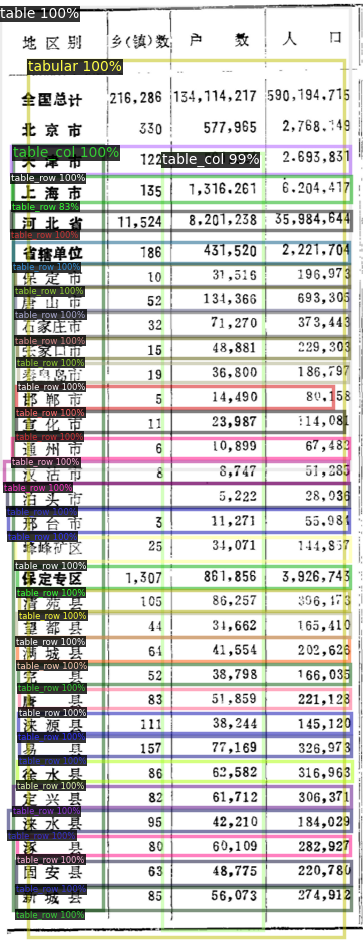}
    \end{tabular}
    \vspace{-1em}
    \caption{A Chinese table parsed by ModernTableParser (left) and HistoricalTableParser (right).}
    \label{fig:M1-vs-M2_ZH}
\end{figure}

In Figures \ref{fig:M1-vs-M2_18-0}, \ref{fig:M1-vs-M2_27-0}, \ref{fig:M1-vs-M2_KO}, and \ref{fig:M1-vs-M2_ZH}, we show the qualitative outputs of ModernTableParser and HistoricalTableParser on various types of inputs.\footnote{Input and annotated figures of original size can be found under \url{https://github.com/DS3Lab/TableParser/tree/main/figures}.} The quality of structure parsing varies across inputs, but overall, the quality is high. Even if we simply use ModernTableParser to parse old scans, it achieves a moderate performance, sometimes better than HistoricalTableParser (see Figures \ref{fig:M1-vs-M2_KO} and \ref{fig:M1-vs-M2_ZH}). This substantiates our claim that knowing the table structure (caption, tabular, row, column, multi-cell, etc.) is of foremost importance for parsing tables. We see that the performance of LayoutParser is quite poor on the tabular data in Figure \ref{fig:HistoricalTableParser} (d) using the best model from its model zoo (PubLayNet/faster\_rcnn\_R\_50\_FPN\_3x).

\section{Related Work}
\paragraph{\textbf{Table Annotation.}}
TableLab \citep{wang2021tablelab} provides an active learning based annotation GUI for users to jointly optimize the model performance under the hood. LayoutParser \citep{shen2021layoutparser} has also promoted an interactive document annotation tool\footnote{See \url{https://github.com/Layout-Parser/annotation-service} (last accessed: Nov. 1, 2021).}, but the tool is not optimized for table annotations. 

\paragraph{\textbf{Table Structure Parsing.}}

As pioneering works in table structure parsing, \cite{kieninger1998t} and \cite{schreiber2017deepdesrt} have both included a review of works in table structure recognition prior to DL. 
Prior methods typically required high human efforts in creating the feature extraction. After \cite{schreiber2017deepdesrt}, researchers have started to revisit table structure parsing with DL methods, which turned out highly promising compared to the rule-based (e.g.,~\cite{kieninger1998t, pivk2007transforming}) and ML-based methods (e.g.,~\cite{pinto2003table,wang2004table, farrukh2017interpreting}). 

The success of DL has marked the revisiting of table structure parsing by \cite{schreiber2017deepdesrt}, which inspired follow-up research  \citep{chi2019complicated, rausch21docparser, prasad2020cascadetabnet, zhong2020image, li2021rethinking, xue2021tgrnet, long2021parsing, jiang2021tabcellnet, nazir2021hybridtabnet, zheng2021global, luo2021deep, raja2020table}. To highlight a few, 
\cite{zhong2020image} proposed EDD (encoder-dual-decoder) to covert table images into HTML code, and they evaluate table recognition (parsing both table structures and cell contents) using a newly devised metric, TEDS (Tree-Edit-Distance-based Similarity). 
\cite{xue2021tgrnet} proposed TGRNet as an effective end-to-end trainable table graph construction network, which encodes a table by combining the cell location detection and cell relation prediction. 
\cite{li2021rethinking} used bi-LSTM on table cell detection by encoding rows/columns in neural networks before the softmax layer. Researchers also started discussing effectively parsing tables in the wild~\citep{long2021parsing}, which is relevant to the perturbation tests we want to conduct for historical tables.
TabCellNet by \cite{jiang2021tabcellnet} adopts a Hybrid Task Cascade network, interweaving object detection and instance segmentation tasks to progressively improve model performance. We see from the previous works, the most effective methods \citep{raja2020table, zheng2021global, jiang2021tabcellnet} always jointly optimize the cell locations and cell relationships. In our work, we consider these two aspects by learning the row and column alignments in a hierarchical structure, where we know the relationship of entities in the table (row, column, cell, caption, footnote).

\section{Discussion and Conclusion}
\subsection{Efficiency}
PyWin32 uses the
component object model (COM), which only supports single-thread processing and only runs under Windows. But with 20 VMs, we managed to process a large amount of files. This is a one-time development cost. On average -- on the fastest machine used (with 16 GB memory, 6 cores, each of 4.8GHz max (2.9 base)) -- it took 15.25 seconds to process one document (a worksheet in this case). To fine-tune a pretrained TableParser with 17 images, it takes 3-4 hours to fine-tune the model with 30k iterations. 

\subsection{Future Work}
Based on our findings, we will further improve the parsing performance on table row/column/cell. Besides, we plan to enable a CSV-export functionality in TableParser, which allows users to export a CSV file that attends to both bounding boxes generated by the OCR'ed and the hierarchical table structure. We will also benchmark this functionality against human efforts. 
Another practical functionality we add to facilitate users' assessment of table parsing quality, is that we enable TableParser to compute row and column sums when exporting to the CSV format. Because tables sometimes come with row/column sums in the rendered format, this functionality can help users to assess their manual efforts in post-editing the CSV output.
We also plan to conduct perturbation tests of table structures and quantify the robustness of our models in those scenarios. These exercises will be highly valuable because, as we see in Figure~\ref{fig:M1-vs-M2_18-0}, we often encounter scan images of tables where the rectangle structures cannot be maintained (the upper right corner). This brings us to another interesting research direction: how to efficiently annotate the non-rectangle elements in a table, e.g., \cite{long2021parsing} have provided the benchmarking dataset and method for parsing tables in the wild. 
Finally, we would like to benchmark TableParser using the popular benchmarking datasets such as ICDAR-2013, ICDAR-2019, TableBank, and PubTabNet. Note that since we develop TableParser on top of the DocParser~\citep{rausch21docparser}, where the reported F1 score has shown superior performance of our method on ICDAR-2013. 

\subsection{Conclusion}
We present in this work our DL-based pipeline to parse table structures and its components: TableAnnotator, TableParser (Modern and Historical), and ExcelAnnotator.
We also demonstrate that pre-training TableParser on weakly annotated data allows highly accurate parsing of structured data in real-world table-form data documents. Fine-tuning the pretrained TableParser in various domains has shown large improvements in detection accuracy. 
We have observed that the state-of-the-art for table extraction is shifting towards DL-based approaches. However, devising suitable tools to facilitate training of such DL approaches for the research community is still lacking. Hence, we provide a pipeline and open-source code and data to invite the active contribution of the community. 

\section{Acknowledgement}

Peter Egger acknowledges Swiss National Science Foundation (Project Number 100018\_204647) for supporting this research project. Ce Zhang and the DS3Lab gratefully acknowledge the support from Swiss National Science Foundation (Project Number 200021\_184628, and 197485), Innosuisse/SNF BRIDGE Discovery (Project Number 40B2-0\_187132), European Union Horizon 2020 Research and Innovation Programme (DAPHNE, 957407), Botnar Research Centre for Child Health, Swiss Data Science Center, Alibaba, Cisco, eBay, Google Focused Research Awards, Kuaishou Inc., Oracle Labs, Zurich Insurance, and the Department of Computer Science at ETH Zurich. Besides, this work would not be possible without our student assistants: We thank Ms.~Ada Langenfeld for assisting us in finding the Hungarian scans and annotating the tables; we thank Mr.~Livio Kaiser for building an ExcelAnnotator prototype during his master thesis. We also appreciate the users' insights on LayoutParser~\citep{shen2021layoutparser} shared by Mr.~Cheongyeon Won. Moreover, the comments and feedback from Sascha Becker and his colleagues at SoDa Labs, Monash University, are valuable in producing the current version of TableParser. We also thank Sascha for providing us with various Korean/European table scans. Finally, we thank the reviewers at \href{https://sites.google.com/view/sdu-aaai22/home}{SDU@AAAI22} for carefully evaluating our manuscripts and their constructive comments. 

\bibliographystyle{aaai}
\bibliography{ref}

\end{document}